\documentclass[letterpaper]{article} 
\usepackage[preprint]{aaai2027}  
\usepackage[hyphens]{url}  
\usepackage{graphicx} 
\usepackage{natbib}  
\usepackage{caption} 
\usepackage{algorithm}
\usepackage{algorithmic}
\usepackage{multirow}
\usepackage{graphicx} 
\usepackage{amsmath}
\usepackage{makecell}
\usepackage{amssymb}

\usepackage{newfloat}
\usepackage{listings}
\DeclareCaptionStyle{ruled}{labelfont=normalfont,labelsep=colon,strut=off} 
\floatstyle{ruled}
\newfloat{listing}{tb}{lst}{}
\floatname{listing}{Listing}

\usepackage{booktabs}

\title{Know Your Step: Faster and Better Alignment for Flow Matching Models via Step-aware Advantages}
\author {
    Zhixiong Yue\textsuperscript{\rm 1, 
    \rm 3}, Feiyang Ye\textsuperscript{\rm 2, \rm 3} \corresponding, Zixuan Ni\textsuperscript{\rm 4}, Sheng Shen\textsuperscript{\rm 4}, Yu Zhang\textsuperscript{\rm 3}
}
\affiliations {
    \textsuperscript{\rm 1}ConrAIn Inc.\\
    \textsuperscript{\rm 2}Li Auto Inc.\\
    \textsuperscript{\rm 3}Southern University of Science and Technology\\
    \textsuperscript{\rm 4}HiThink Research\\
}

\begin{document}

\maketitle

\begin{abstract}
Recent advances in flow matching models, particularly with reinforcement learning (RL), have significantly enhanced human preference alignment in few-step text-to-image generators. However, existing RL-based  approaches for flow matching models typically rely on numerous denoising steps, while suffering from sparse and imprecise reward signals that often lead to suboptimal alignment.
To address these limitations, we propose Temperature-Annealed Few-step Sampling with Group Relative Policy Optimization (TAFS-GRPO), a novel framework for training flow matching text-to-image models into efficient few-step generators well aligned with human preferences. 
Our method iteratively injects adaptive time-dependent noise into one-step clean predictions. By repeatedly annealing the model’s sampled outputs, it introduces stochasticity into the sampling process while preserving the semantic integrity of each generated image. 
Moreover, its step-aware advantage integration mechanism combines GRPO with temperature-annealed sampling to eliminate the need for a differentiable reward function and provide dense, step-specific rewards for stable policy optimization.
Extensive experiments demonstrate that TAFS-GRPO achieves strong performance in few-step text-to-image generation and significantly improves the alignment of generated images with human preferences. The code and models of this work will be available to facilitate further research.
\end{abstract}

\section{Introduction}
\label{sec:intro}

Flow matching models have recently emerged as a popular paradigm for Text-to-Image (T2I) generation, acclaimed for their simplicity and ability to produce high-quality images \cite{blackforestlabs2024flux, esser2024scaling, cai2025hidream, xie2025sana}. However, the significant inference latency, coupled with the escalating computational costs that grow with the scale of inference steps, presents a major obstacle to its effective application in real-world business scenarios.
While techniques like quantization can accelerate inference, they often lead to a significant degradation in generation quality \cite{li2024svdquant,zhao2024vidit,liu2024hq}. Recent advances in few-step distillation, especially reinforcement learning (RL) based approaches, aim to preserve quality while drastically reducing sampling steps \cite{clark2023directly, xu2023imagereward, luo2024diff, luo2025reward}.

\begin{figure}[t]
\centering

\includegraphics[width=\linewidth]{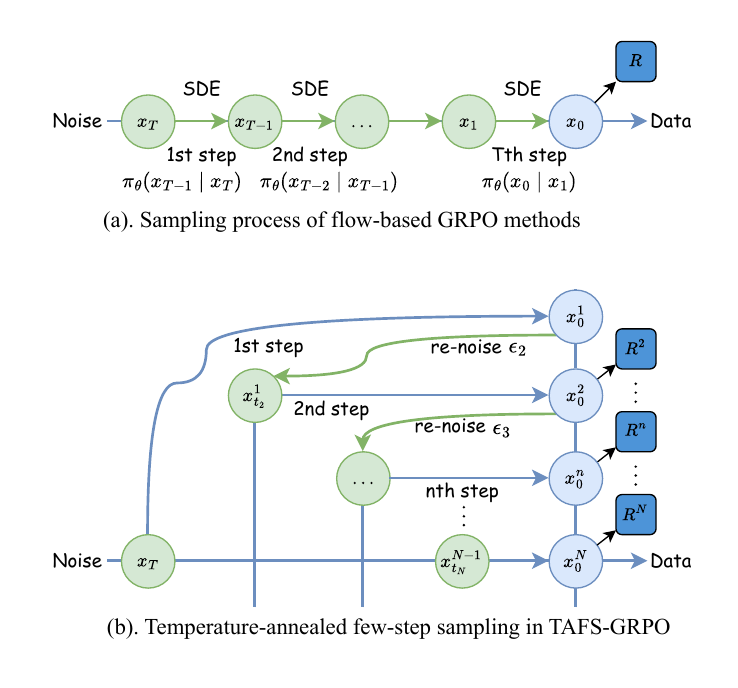}

\caption{Comparison of the sampling processes of flow-based GRPO methods and TAFS-GRPO.}
\label{fig:mcs}

\end{figure}

\begin{figure*}[t]
\centering
\includegraphics[width=0.92\linewidth]{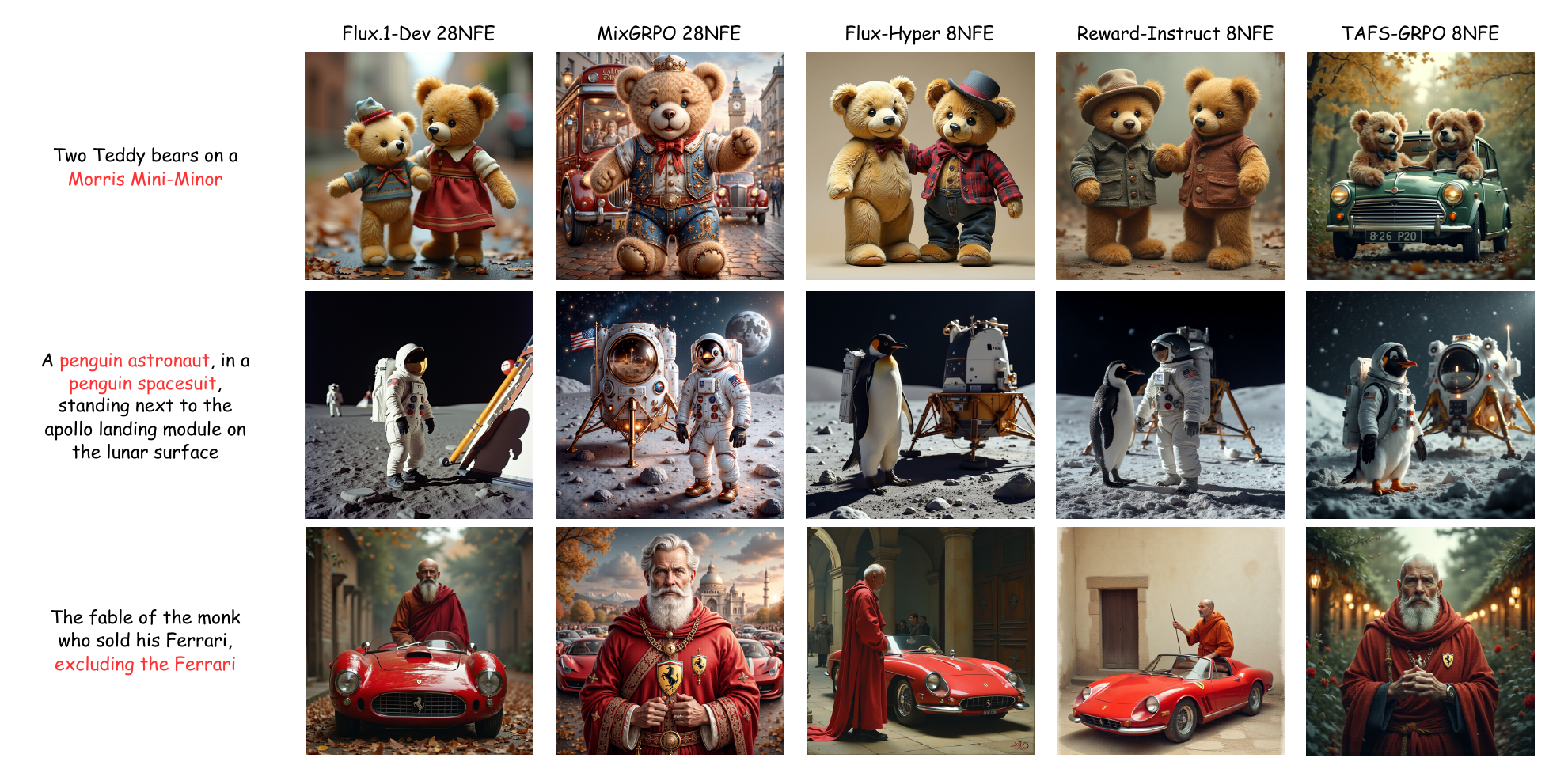}
\caption{Qualitative comparison of generations from 28 sampling steps of Flux.1-dev, MixGRPO, and 8 sampling steps of Hyper-Flux, Reward-Instruct, and the proposed TAFS-GRPO. TAFS-GRPO demonstrates superior performance in semantic and aesthetic alignment with fewer sampling steps than the baseline Flux.1-dev.} 
\label{fig:visual}
\end{figure*}

RL-based training approaches rely on efficient sampling to collect rewards, but flow matching models typically require numerous iterative denoising steps (e.g., 20 to 40 steps or more) to generate a single sample \cite{liu2025flow, xue2025dancegrpo}. 
This gives rise to two primary challenges. 

First, the sequential nature of the generation process creates a significant computational bottleneck, severely limiting training efficiency and scalability \cite{li2025mixgrpo, yu2025smart}. 
This inefficiency is particularly pronounced in online RL settings where numerous samples must be generated and evaluated in real-time during training \cite{miao2024tuning, oertell2024rl}. 
For example, RL-guided distillation methods \cite{li2024reward, ren2024hyper, luo2024diff} demand full-step sampling of the flow models to calculate the diffusion distillation loss while applying reward-guided distillation.
Although reward-centric methods \cite{clark2023directly, luo2025reward, jia2025reward} eliminate the need for an expensive diffusion-distillation loss, they remain dependent on the complete denoising trajectory for reward computation.
Furthermore, these methods rely on differentiable reward functions, limiting their applicability.

Second, the large number of denoising steps of flow matching models leads to a sparse reward problem, as a meaningful reward signal is only available after the complete denoising trajectory \cite{hu2025towards, zhou2025text}.
This sparsity makes it exceedingly difficult to discern which specific denoising step taken during the generation process positively contributes to the final alignment outcome. 
Consequently, policies across all timesteps receive the same terminal reward \cite{liu2025flow, xue2025dancegrpo}.
This uniform credit assignment based on sparse terminal rewards fails to accurately credit contributions of different denoising steps, resulting in inefficient exploration and suboptimal convergence \cite{he2025tempflow}.
These challenges underscore the need for continued research into novel RL distillation frameworks that can overcome the sampling inefficiency and credit assignment problems inherent in fine-tuning flow matching models for enhanced human preference alignment.

To jointly address these two challenges, this work introduces Temperature-Annealed Few-step Sampling with Group Relative Policy Optimization (TAFS-GRPO), which addresses the sampling inefficiency and sparse reward problem commonly observed in flow matching RL-based distillation frameworks. It contains two key components: \textbf{Temperature-Annealed Sampling} and \textbf{Step-aware Advantage Integration}. The temperature-annealed sampling combines one-step sampling with time-dependent noise injection. By adding adaptive noise at different temporal steps to the result of a one-step sample, the procedure repeatedly anneals the sampling outcome. This iterative rollback mechanism not only introduces stochasticity into the sampling process but also produces the essential semantic information at each sample, which significantly alleviates the reward sparsity problem commonly observed in flow matching models.
The step-aware advantage integration combines the GRPO-based RL strategy with temperature-annealed sampling outcomes. While avoiding the need for the reward function to be differentiable, this module evaluates the image produced at each annealing step within every sample trajectory. By assigning an advantage score to each sample output, the method substantially increases the amount of informative reward available at each sampling step and mitigates the negative effects of sparse rewards on GRPO. Because of this, TAFS-GRPO not only increases the sampling frequency achievable by RL-based methods on flow matching models and alleviates reward sparsity, but also stabilizes policy optimization in the few-step generation regime. The experiments demonstrate that, under the same training configuration, our method achieves faster convergence compared to Reward-Instruct \cite{luo2025reward}, and consistently outperforms it across all image quality metrics at both 4-step and 8-step sampling. The contributions of this paper are summarized as follows:

\begin{itemize}
\item We first introduce temperature-annealed sampling to the GRPO training framework for flow matching models, leveraging the noise addition process to provide a stochastic environment for online RL.

\item We propose a novel RL-based framework named TAFS-GRPO, which improves the sampling efficiency and addresses the sparse reward problem by the step-aware advantage integration mechanism while training.

\item The TAFS-GRPO trained model achieves high-quality results with far fewer steps, significantly speeding up inference. The model maintains high performance across a wide range of sampling steps, providing flexibility for deployment under different latency constraints.
\end{itemize}

\section{Related Work}
\label{sec:related-work}

\subsection{Preference Alignment for T2I Models}
The alignment of T2I models with human preferences has become a critical research direction to enhance the usability, safety, and aesthetic quality of generated content. A prominent line of work adapts Reinforcement Learning from Human Feedback (RLHF) for diffusion models. Early preference-optimization approaches such as Direct Preference Optimization (DPO) \cite{wallace2024diffusion} and its variants \cite{wang2025diffusion,liang2025aesthetic,zhang2025diffusion} directly optimize denoising policies from pairwise preference data without explicit on-policy policy-gradient updates. These methods often suffer from high computational cost and instability. 

To address these limitations, group-based RL methods such as GRPO have been proposed to improve sample efficiency by computing relative advantages over a group of trajectories sampled independently for the same prompt.
DanceGRPO \cite{xue2025dancegrpo} is highlighted as the first unified framework to adapt GRPO to visual generation, enabling its application across diverse generative paradigms (diffusion models and rectified flows), tasks (text-to-image, text-to-video, image-to-video), foundation models, and reward models. It leverages Stochastic Differential Equations (SDE) sampling to introduce randomness and addresses training instability, demonstrating substantial improvements on the human preference alignment task. Flow-GRPO \cite{liu2025flow} integrates online RL into flow matching models by recasting the original deterministic Ordinary Differential Equation (ODE) as an equivalent SDE, formulating the denoising process as a Markov decision process. 

To mitigate the computational overhead of full-step sampling in these methods, MixGRPO \cite{li2025mixgrpo} proposes a mixed ODE-SDE sampling strategy with a sliding window mechanism. TempFlow-GRPO \cite{he2025tempflow} addresses the temporal uniformity assumption in previous GRPO methods by introducing a trajectory branching strategy for precise credit assignment, leading to more temporally-aware optimization. DenseGRPO \cite{deng2026densegrpo} obtains dense rewards by completing intermediate noisy states, whereas TAFS changes the rollout construction so that each stochastic transition is followed by a reusable clean prediction. Pref-GRPO \cite{wang2025pref} fundamentally reformulates the optimization objective from absolute reward score maximization to pairwise preference fitting. 

However, existing preference alignment methods have paid little attention to the few-step generation setting, leaving its training paradigm largely unexplored.

\subsection{Few-step Distillation of T2I models}

Recent advances in T2I model distillation have increasingly adopted trajectory-based methods to enable efficient few-step sampling. Methods such as Latent Consistency Models (LCM) \cite{luo2023latent} and SANA-Sprint \cite{chen2025sana} learn the solution trajectory of the probability flow ODE (PF-ODE) over reduced time intervals, enforcing self-consistency across steps to approximate the original denoising process with minimal inference steps. 

In contrast, distribution-based distillation aligns the student model’s generative distribution with that of the teacher model. This category includes GAN-based approaches like LADD \cite{sauer2024fast}, which employs adversarial training in latent space, and score distillation methods such as DMD \cite{yin2024one} and its successor DMD2, which minimize distribution divergence without relying on instance-level trajectory matching.

Despite their effectiveness, these distillation techniques are often computationally intensive and typically depend on real image datasets. Recent efforts such as RG-LCD \cite{li2024reward} and DI++ \cite{luo2024diff} incorporate reward maximization into the distillation process. However, they require training an additional score model to maintain proximity to the original generator, incurring substantial memory and computational overhead. LaSRO \cite{jia2025reward} addresses this by leveraging latent space surrogate rewards to optimize arbitrary reward signals through efficient off-policy exploration. Similarly, Reward-Instruct \cite{luo2025reward} observes that reward gradients dominate training, reducing diffusion distillation to a costly regularization role. It distills pre-trained diffusion models into reward-aligned few-step generators without distillation losses or training images. Nevertheless, this reward-centric approach still relies on differentiable reward functions, limiting its applicability and introducing computational inefficiency.

In comparison, TAFS-GRPO eliminates the need for differentiable reward functions by leveraging a policy gradient algorithm. Moreover, it introduces a step-aware advantage integration mechanism to provide precise and dense evaluation feedback for the outcome of each stochastic sampling action, which addresses the sparse reward problem.

\begin{figure*}[!htbp]
\centering
\includegraphics[width=0.98\linewidth]{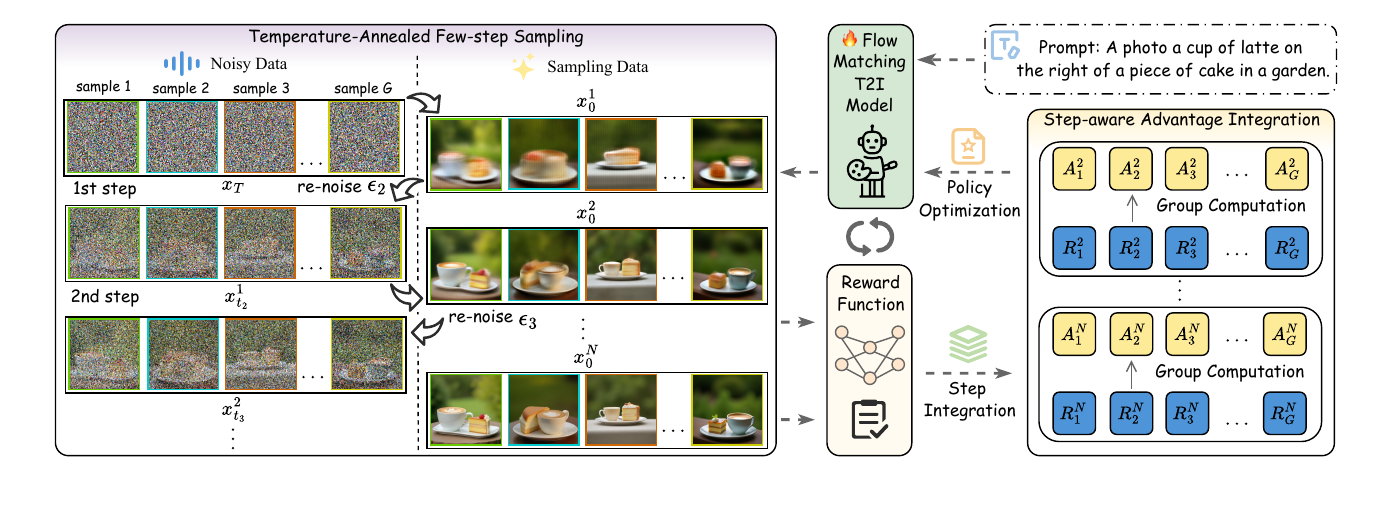}
\caption{The architecture of the proposed TAFS-GRPO framework. Given a dataset of prompts, our temperature-annealed few-step sampling generates a clean estimate at every sampling step. The step-aware advantage integration mechanism evaluates the clean estimates from the second prediction onward and computes stage-wise group-relative advantages. These precise feedback signals guide the few-step model updates.}
\label{fig:framework}
\end{figure*}

\section{Method} \label{sec:method}
We review GRPO and its compatibility with non-differentiable rewards, and then discuss the sparse rewards in conventional flow-based GRPO. Finally, we introduce TAFS-GRPO, which combines \textbf{Temperature-Annealed Sampling} with \textbf{Step-aware Advantage Integration} to enable efficient few-step policy optimization.

\subsection{GRPO with Non-Differentiable Rewards}
Group Relative Policy Optimization (GRPO) estimates the advantage of each sampled output relative to other outputs generated for the same condition \cite{guo2025deepseek,shao2024deepseekmath}. Given a condition $q$, a group of outputs $\{o_i\}_{i=1}^{G}$ is sampled from the old policy $\pi_{\theta_{\mathrm{old}}}$. The general clipped GRPO objective is
\begin{equation}
\footnotesize
\begin{split}
\mathcal{J}_{\mathrm{GRPO}}(\theta)
&=\mathbb{E}\Bigg[
\frac{1}{G}\sum_{i=1}^{G}\frac{1}{|o_i|}
\sum_{t=1}^{|o_i|}
\min\big(
\rho_{i,t}(\theta)\hat{A}_{i,t},\\
&\operatorname{clip}(\rho_{i,t}(\theta),1-\varepsilon,1+\varepsilon)
\hat{A}_{i,t}\big)\Bigg]
-\beta\mathbb{D}_{\mathrm{KL}}[\pi_\theta\Vert\pi_{\mathrm{ref}}]
,
\end{split}
\label{eq:GRPO-obj}
\end{equation}
where
\begin{equation}
\rho_{i,t}(\theta)=
\frac{\pi_\theta(o_{i,t}\mid q,o_{i,<t})}
{\pi_{\theta_{\mathrm{old}}}(o_{i,t}\mid q,o_{i,<t})}
\label{eq:GRPO-ratio}
\end{equation}
is the importance sampling ratio and $\hat{A}_{i,t}$ is the group-relative advantage. A typical outcome-level advantage is obtained by normalizing rewards within the group:
\begin{equation}
\hat{A}_i=\frac{R(o_i,q)-\operatorname{mean}(\mathbf{R})}
{\operatorname{std}(\mathbf{R})}.
\end{equation}
The reward is treated as a scalar with stopped gradients when optimizing Eq.~(\ref{eq:GRPO-obj}). Consequently, GRPO does not require the reward function itself to be continuous or differentiable, allowing arbitrary task-specific evaluators to provide policy feedback.

\subsection{Sparse Rewards in Flow-Based GRPO}
Flow matching models commonly generate samples by solving an ODE. Conditioned on an initial noise $x_T$, an ODE trajectory is deterministic and therefore does not provide the non-degenerate transition density required to compute step-wise policy likelihood ratios. Existing flow-based GRPO methods address this issue by converting the ODE sampler into an equivalent stochastic sampler, typically based on an SDE. The resulting stochastic transitions make policy optimization possible.

However, an intermediate noisy state $x_t$ generally does not contain sufficient clean visual information for an image-space reward model. As illustrated in Figure~\ref{fig:mcs}(a), the reward can therefore be reliably evaluated only on the final clean output $x_0$. All transitions along a long sampling trajectory consequently share the same terminal reward, making the feedback sparse and temporally coarse. Moreover, obtaining each reward requires executing the complete multi-step trajectory, which substantially increases the sampling cost of online GRPO training.

\subsection{Temperature-Annealed Few-Step Sampling}
TAFS-GRPO replaces a long stochastic denoising trajectory with a sequence of one-step ODE predictions connected by temperature-controlled re-noising transitions. The first one-step ODE prediction is deterministic conditioned on its initial noise. Each subsequent prediction is stochastic because its input is sampled by re-noising the preceding clean prediction. This construction yields a semantically meaningful clean estimate after every annealing transition, enabling rewards to be evaluated from the second clean estimate onward.

\noindent\textbf{Temperature-Annealed Sampling}.
Let $v_\theta(x_t,t,\mathbf{c})$ denote the conditional velocity field. Under our time convention, the one-step clean prediction from state $x_t$ is
\begin{equation}
F_\theta(x_t,t,\mathbf{c})
=x_t+t\,v_\theta(x_t,t,\mathbf{c}).
\label{eq:one-step-prediction}
\end{equation}
We use $N$ one-step clean predictions in total and connect them with $N-1$ annealing transitions. The corresponding time interval is $\tau=T/N$, and the time assigned to the $n$-th prediction is $t_n=T-(n-1)\tau$. We interpret the scheduler noise level $\sigma_{t_n}$ as the stage-wise temperature, which decreases along the annealing trajectory.

Starting from $x_T\sim\mathcal{N}(0,I)$, the first clean estimate is
\begin{equation}
x_0^1=F_\theta(x_T,T,\mathbf{c}).
\label{eq:first-clean}
\end{equation}
Conditioned on $x_T$, this prediction is deterministic. We therefore use it to initialize the annealing trajectory but do not assign it a step reward.

For each subsequent prediction $n=2,\ldots,N$, we re-noise the preceding clean estimate according to the flow model's noise schedule:
\begin{equation}
x_{t_n}^{n-1}
=\alpha_{t_n}x_0^{n-1}
+\sigma_{t_n}\epsilon_n,
\qquad
\epsilon_n\sim\mathcal{N}(0,I),
\label{eq:annealed-transition}
\end{equation}
where $\alpha_{t_n}$ and $\sigma_{t_n}$ are determined by the scheduler \cite{blackforestlabs2024flux, esser2024scaling}. A one-step ODE prediction from this random state then produces
\begin{equation}
x_0^n=F_\theta(x_{t_n}^{n-1},t_n,\mathbf{c}).
\label{eq:annealed-clean}
\end{equation}
Thus, although Eq.~(\ref{eq:annealed-clean}) is deterministic conditioned on $x_{t_n}^{n-1}$, the clean estimate $x_0^n$ is stochastic due to the noise $\epsilon_n$ injected in Eq.~(\ref{eq:annealed-transition}). Repeating the re-noise--predict procedure progressively reduces the noise level while retaining a reward-compatible clean estimate at every stage.

For analysis, we embed the training-time full-refresh transition
into a stochasticity-interpolated family and extend the finite
training schedule to the clean endpoint, see
Appendix. Under an oracle endpoint coupling,
this analytical family preserves the corresponding per-time
rectified-flow bridge marginals, while its zero-stochasticity limit
recovers the deterministic explicit-Euler ODE update.

\noindent\textbf{Transition Likelihood}.
During trajectory collection, all clean estimates and annealed states are generated with the old policy $\pi_{\theta_{\mathrm{old}}}$. Let $s_i^{n-1}$ denote the fixed input state used to produce $x_{0,i}^{n-1}$ in the collected trajectory. Because the mean of the subsequent re-noising transition depends on this policy prediction, its conditional density under a candidate policy $\pi_\theta$ is tractable:
\begin{equation}
\begin{split}
p_\theta(x_{t_n,i}^{n-1}\mid s_i^{n-1},\mathbf{c})
=\mathcal{N}\Big(
x_{t_n,i}^{n-1};
\,\alpha_{t_n}F_\theta(s_i^{n-1},\mathbf{c}),
\sigma_{t_n}^{2}I
\Big).
\end{split}
\label{eq:transition-density}
\end{equation}
Here, $F_\theta(s_i^{n-1},\mathbf{c})$ is shorthand for Eq.~(\ref{eq:one-step-prediction}) evaluated at the latent and time contained in $s_i^{n-1}$. Equation~(\ref{eq:transition-density}), rather than the deterministic conditional mapping in Eq.~(\ref{eq:annealed-clean}), defines the policy likelihood used by GRPO.

\noindent\textbf{Step-aware Advantage Integration}.
For each prompt, we sample a group of $G$ trajectories. At every stage $n\geq2$, the clean estimates generated after the same number of annealing transitions are evaluated together:
\begin{equation}
R_i^n=\hat{R}(x_{0,i}^n,\mathbf{c}),\qquad n=2,\ldots,N.
\label{eq:step-reward}
\end{equation}
We then normalize the rewards across the group at each stage,
$A_i^n=(R_i^n-\mu_n) / s_n$,
where $\mu_n$ and $s_n$ are respectively the mean and standard deviation of $\{R_i^n\}_{i=1}^{G}$. This stage-wise normalization compares samples at the same noise level and provides dense, step-specific feedback.

The importance sampling ratio associated with $R_i^n$ is computed on the realized annealing transition that leads to the $n$-th clean estimate:
\begin{equation}
\rho_i^n(\theta)=
\frac{p_\theta(x_{t_n,i}^{n-1}\mid s_i^{n-1},\mathbf{c})}
{p_{\theta_{\mathrm{old}}}(x_{t_n,i}^{n-1}\mid s_i^{n-1},\mathbf{c})}.
\label{eq:tafs-ratio}
\end{equation}
Rather than summing the advantages into a single trajectory-level value, we form a clipped GRPO surrogate for every valid stage and average these step-specific objectives:
\begin{equation}
\begin{split}
\mathcal{J}_{\mathrm{policy}}(\theta)
=&\frac{1}{G(N-1)}
\sum_{i=1}^{G}\sum_{n=2}^{N}
\Big[
\min\Big(\rho_i^n(\theta)A_i^n,\\
&\operatorname{clip}(\rho_i^n(\theta),1-\varepsilon,1+\varepsilon)A_i^n
\Big)\\
&-\beta\mathbb{D}_{\mathrm{KL}}\!\left[
p_\theta(\cdot\mid s_i^{n-1},\mathbf{c})
\Vert
p_{\mathrm{ref}}(\cdot\mid s_i^{n-1},\mathbf{c})
\right]
\Big].
\end{split}
\label{eq:tafs-objective}
\end{equation}
Here, $p_{\mathrm{ref}}$ is the frozen transition policy induced by the reference flow model. The old policy $p_{\theta_{\mathrm{old}}}$ is used to form the importance ratio, whereas $p_{\mathrm{ref}}$ prevents the optimized policy from drifting excessively from the reference model. The reward values and advantages are detached during policy optimization. Consequently, Eq.~(\ref{eq:tafs-objective}) supports non-differentiable reward functions while integrating feedback from all stochastic annealing stages.

\begin{algorithm}[t]
\caption{TAFS-GRPO Training Process}
\label{alg:TAFS-GRPO}

\textbf{Input}: Prompt dataset $\mathcal{C}$, policy $\pi_\theta$, reward function $\hat{R}$, time horizon $T$, number of clean predictions $N$, group size $G$\\
\textbf{Output}: Few-step policy $\pi_\theta^*$

\begin{algorithmic}[1]
\FOR{training iteration $e=1$ to $E$}
    \STATE Set old policy $\pi_{\theta_{\mathrm{old}}}\leftarrow\pi_\theta$
    \STATE Sample a prompt batch $\mathcal{C}_b\sim\mathcal{C}$
    \FOR{prompt $\mathbf{c}\in\mathcal{C}_b$}
        \STATE Sample $\{x_{T,i}\}_{i=1}^{G}\sim\mathcal{N}(0,I)$
        \STATE $x_{0,i}^{1}\leftarrow F_{\theta_{\mathrm{old}}}(x_{T,i},T,\mathbf{c})$ for all $i$
        \FOR{$n=2$ to $N$}
            \STATE $t_n\leftarrow T-(n-1)T/N$
            \STATE Sample $\epsilon_{n,i}\sim\mathcal{N}(0,I)$ for all $i$
            \STATE $x_{t_n,i}^{n-1}\leftarrow\alpha_{t_n}x_{0,i}^{n-1}+\sigma_{t_n}\epsilon_{n,i}$
            \STATE $x_{0,i}^{n}\leftarrow F_{\theta_{\mathrm{old}}}(x_{t_n,i}^{n-1},t_n,\mathbf{c})$ for all $i$
            \STATE $R_i^n\leftarrow\hat{R}(x_{0,i}^{n},\mathbf{c})$ for all $i$
            \STATE Compute $\mu_n$, $s_n$, and $A_i^n\leftarrow(R_i^n-\mu_n)/s_n$
            \STATE Store the realized transition and its old-policy log probability
        \ENDFOR
    \ENDFOR
    \STATE Compute $\rho_i^n(\theta)$ on the stored transitions using Eq.~(\ref{eq:tafs-ratio})
    \STATE Compute $\mathcal{J}_{\mathrm{policy}}(\theta)$ using Eq.~(\ref{eq:tafs-objective})
    \STATE Update $\theta$ by gradient ascent on $\mathcal{J}_{\mathrm{policy}}(\theta)$
\ENDFOR
\STATE \textbf{return} $\pi_\theta^*$
\end{algorithmic}
\end{algorithm}

As summarized in Algorithm~\ref{alg:TAFS-GRPO}, an $N$-prediction trajectory contains one deterministic initialization and $N-1$ stochastic annealing transitions. It requires $N$ old-policy evaluations to collect the trajectory and $N-1$ policy evaluations to recompute the transition likelihood ratios. Rewards are evaluated only for $x_0^2,\ldots,x_0^N$, and their corresponding clipped GRPO objectives are averaged rather than first collapsing the step-wise advantages into a single trajectory-level advantage.

\begin{table*}[t]
\centering
\resizebox{\linewidth}{!}{
\begin{tabular}{c|lccccccc}
\toprule
& Methods  & NFE $\downarrow$ & GenEval $\uparrow$ & Pick Score $\uparrow$ 
& CLIP $\uparrow$ 
& HPS-v2.1 $\uparrow$ & ImageR. $\uparrow$ & Unified R. $\uparrow$ \\
\midrule
\multirow{6}{*}{\rotatebox{90}{Pre-train}}
& SD3.5-M \cite{esser2024scaling} & 40 & 67.99 & 22.95 & 28.70 & 0.285 & 0.947 & 3.496 \\
& SD3.5-Large \cite{esser2024scaling} & 28 & 73.80 & 23.21 & 29.08 & 0.293 & 1.051 & 3.622 \\
& Flux.1-dev \cite{blackforestlabs2024flux} & 28 & 67.54 & 23.41 & 28.19 & 0.305 & 1.013 & 3.562 \\
& SANA-1.5 1.6B \cite{xie2025sana} & 20 & 62.48 & 23.06 & 28.86 & 0.301 & 0.969 & 3.412 \\
& HiDream-I1-Full \cite{cai2025hidream} & 50 & 80.46 & 23.47 & 28.84 & 0.325 & 1.334 & 3.748 \\
& Flux.2-klein-4B-base \cite{blackforestlabs2026flux2kleinbase4b} & 50 & 76.83 & 18.99 & 29.39 & 0.295 & 1.060 & 3.765 \\
\cmidrule{1-9}
\multirow{12}{*}{\rotatebox{90}{Step-distillation}}
& SD3.5-Large-Turbo \cite{esser2024scaling} & 4 & 69.62 & 23.18 & 28.90 & 0.287 & 0.927 & 3.522 \\
& Flux.1-schnell \cite{blackforestlabs2024flux} & 4 & 67.82 & 23.01 & 28.61 & 0.295 & 0.918 & 3.538 \\
& Flux-Turbo-Alpha \cite{flux-turbo} & 4 & 61.41 & 23.06 & 27.57 & 0.290 & 0.837 & 3.404 \\
& Hyper-Flux \cite{ren2024hyper} & 8 & 70.03 & \underline{23.55} & 28.16 & 0.311 & 1.005 & 3.558 \\
& SANA-Sprint 1.6B \cite{chen2025sana} & 4 & 71.33 & 23.13 & 28.81 & 0.307 & 1.130 & 3.546 \\
& HiDream-I1-Fast \cite{cai2025hidream} & 16 & 76.89 & 23.53 & 28.53 & 0.305 & 1.298 & 3.749 \\
& Flux.2-klein-4B \cite{blackforestlabs2026flux2klein4b} & 4 & 76.34 & 18.96 & 29.31 & 0.304 & 1.221 & \underline{3.878} \\
\cmidrule{2-9} 
& TAFS-GRPO+SD3.5-M & 4 & 80.84 & 22.61 & \textbf{30.03} & 0.274 & 1.052 & 3.598 \\
& TAFS-GRPO+SD3.5-M & 8 & 82.37 & 23.03 & 29.87 & 0.287 & 1.170 & 3.683 \\
& TAFS-GRPO+Flux.1-dev & 4 & \underline{84.75} & 23.39 & \underline{29.93} & 0.295 & \underline{1.437} & 3.809 \\
& TAFS-GRPO+Flux.1-dev & 8 & \textbf{86.52} & \textbf{23.67} & 29.62 & \textbf{0.330} & \textbf{1.448} & 3.857 \\
& TAFS-GRPO+Flux.2-klein-4B & 4 & 78.16 & 19.42 & 28.97 & \underline{0.329} & 1.392 & \textbf{3.958} \\
\bottomrule
\end{tabular}
}
\caption{Comprehensive comparison of TAFS-GRPO with step distillation approaches in performance on the composition image generation task. ImageR. and Unified R. stand for ImageReward and Unified Reward. $\uparrow$ ($\downarrow$) indicates the higher (lower) the result, the better the performance. We highlight the best and second best entries.}
\label{tab:distill}
\end{table*}

\section{Experiments}\label{sec:exp}
We conduct our experiments on three flow matching T2I models: SD3.5-M \cite{esser2024scaling}, Flux.1-dev \cite{blackforestlabs2024flux}, and the four-step distilled Flux.2-klein-4B \cite{blackforestlabs2026flux2klein4b}. We consider the composition image generation task GenEval \cite{ghosh2023geneval} and the human preference alignment task Pick-a-Pic \cite{kirstain2023pick}. The composition image generation task assesses T2I models on complex compositional prompts, and the human preference alignment task aims to align T2I models with human preferences. These tasks empirically evaluate the ability of TAFS-GRPO to distill and improve flow matching models across different model families and initial sampling budgets.

\subsection{Experimental Setup}
We introduce two tasks and elaborate on their base models, training prompts, and reward metrics.
For both tasks, we applied the Low-Rank Adaptation (LoRA) \cite{hu2022lora} method for text to image generation, following \cite{xue2025dancegrpo, li2025mixgrpo, liu2025flow}.
For the composition image generation task, we employ TAFS-GRPO on SD3.5-M, Flux.1-dev, and Flux.2-klein-4B with the prompt dataset provided by GenEval, following \cite{liu2025flow, xue2025advantage}. The SD3.5-M, Flux.1-dev and Flux.2-klein-4B settings are detailed in the Appendix.
For the human preference alignment task, we use Flux.1-dev as the base model. We perform experiments using the prompt provided by the Pick-a-Pic dataset \cite{kirstain2023pick} with 25432 text prompts for training and 2048 diverse text prompts for testing as suggested in \cite{liu2025flow, xue2025advantage}. All methods are evaluated with 8 inference steps to ensure fairness. \textit{Due to the page limitation, the implementation details and detailed evaluation metrics are put in Appendix.}

\begin{table*}[htbp]
\centering
\resizebox{\linewidth}{!}{
\begin{tabular}{lccccccccc}
\toprule
\multirow{2}{*}{Method} & \multirow{2}{*}{\makecell{NFE$_{\pi_{\theta_{\text{old}}}}$ \\ / NFE$_{\pi_{\theta}}$}} 
& \multirow{2}{*}{\makecell{Iteration \\ Time (s) $\downarrow$}} 
& \multicolumn{2}{c}{In-Domain} & \multicolumn{3}{c}{Out-of-Domain} \\
\cmidrule(lr){4-5} \cmidrule(lr){6-8}
& & &  Pick Score $\uparrow$ & CLIP Score $\uparrow$ & HPS-v2.1 $\uparrow$ & ImageR. $\uparrow$ & Unified R. $\uparrow$ \\
\midrule
Flux.1-dev \cite{blackforestlabs2024flux} & - & -  & 21.68 & 25.95 & 0.280 & 0.848 & 3.328 \\
\midrule
DanceGRPO \cite{xue2025dancegrpo} & 25 / 14 & 646 & 22.20 & 27.42 & 0.333 & 1.212 & 3.484 \\
MixGRPO \cite{li2025mixgrpo} & 25 / 4 & 334 & 22.23 & 27.67 & 0.324 & 1.210 & 3.472 \\
Flow-GRPO \cite{liu2025flow} & 10 / 6 & 248 & 22.26 & 27.72 & 0.304 & 1.035 & 3.460  \\
DenseGRPO \cite{deng2026densegrpo} & 4 / 3 & 225 & 22.09 & \textbf{27.92} & 0.297 & 1.017 & 3.500 \\
\midrule
RG-LCD \cite{li2024reward} & - & 466 & 21.97 & 26.98 & 0.283 & 0.929 & 3.336 \\
Reward-Instruct \cite{luo2025reward} & - & 206 & 22.13 & 27.23 & 0.286 & 0.973 & 3.392 \\
TAFS-GRPO & 4 / 3 & \textbf{116} & \textbf{22.46} & 27.68 & \textbf{0.353} & \textbf{1.595} & \textbf{3.511} \\
\bottomrule
\end{tabular}
}
\caption{
Comparison of sampling efficiency and in-domain and out-of-domain performance on the human preference alignment task. ImageR. and Unified R. stand for ImageReward and Unified Reward. $\uparrow$ ($\downarrow$) indicates that the higher (lower) the result, the better the performance. The number of Function Evaluations (NFE) and the time consumption per iteration with the same number of samples are used for the evaluation of computational overhead.
NFE$_{\pi_{\theta_{\text{old}}}}$ and NFE$_{\pi_{\theta}}$ denote the numbers of forward passes through the old and updated policies, respectively, when computing the policy ratio. We highlight the best entries.
}
\label{tab:quantitative}
\end{table*}

\subsection{Main Results}

\begin{figure}[t]
\centering
\includegraphics[width=0.85\linewidth]{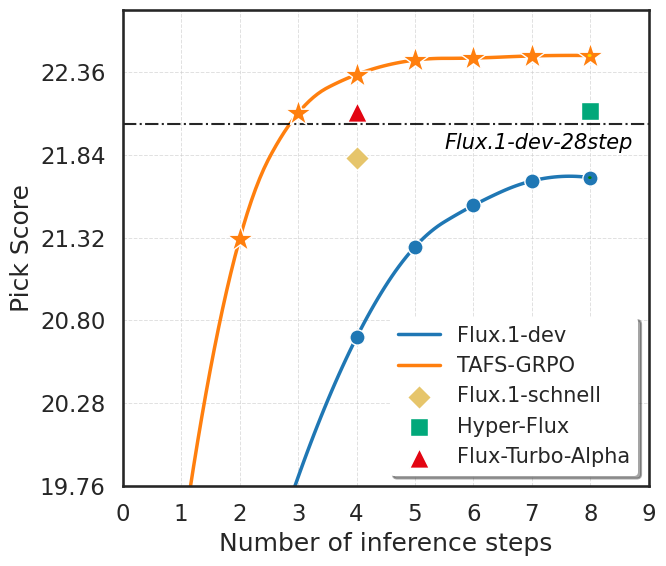}
\caption{
The pick score of various models based on Flux.1-dev at different numbers of inference steps. TAFS-GRPO sustains high performance across 4 to 8 inference steps, illustrating its flexibility under different computational budgets.
}
\label{fig:step_pickscore}
\end{figure}

\noindent\textbf{Composition Image Generation.}
Table~\ref{tab:distill} shows the results of TAFS-GRPO on the composition image generation task compared with various step distillation methods. 
Our TAFS-GRPO+Flux model not only surpasses other Flux-based step-distilled models, like Hyper-Flux (RL-guided step-distilled model) and Flux-Turbo-Alpha (distilled with extra data) on all metrics, but also outperforms larger models such as SD3.5-Large and HiDream-I1-Full.
Our comparative study does not include DanceGRPO, MixGRPO, and Flow-GRPO because these methods do not employ step distillation techniques. We excluded RG-LCD and Reward-Instruct from our comparisons because these methods require the reward model to be differentiable. Our method achieves competitive results on both SD3.5-M and Flux, demonstrating its strong generalizability across different base models.

\noindent\textbf{Human Preference Alignment.}
For the human preference alignment task, we evaluated the computational overhead and performance of TAFS-GRPO compared to flow-based GRPO methods \cite{liu2025flow,li2025mixgrpo,xue2025dancegrpo} and RL-based distillation methods \cite{luo2025reward,li2024reward}, with the results presented in Table~\ref{tab:quantitative}.
Compared to Flow-GRPO, TAFS-GRPO achieves significant improvements in various out-of-domain metrics under the same number of inference steps, while also achieving a 2.14$\times$ speedup. This demonstrates that temperature-annealed few-step sampling effectively enhances training efficiency. Under the same 4-step rollout horizon and the same 3-step policy replay budget, TAFS-GRPO is faster end-to-end because it reuses the clean estimates produced during the rollout, whereas DenseGRPO performs 10 additional ODE-completion model evaluations per sample to construct state-wise rewards. Due to the expensive distillation loss, RG-LCD trains significantly slower than TAFS-GRPO and Reward-Instruct.
TAFS-GRPO achieves a relative improvement of 24.7\% on HPS-v2.1 compared to RG-LCD, achieving superior alignment with human preferences.
This indicates that step-aware advantage integration yields precise reward signals to enhance the optimization of the distillation model. Our approach delivers robust performance on the tasks of compositional image generation and human preference alignment, showing that it generalizes effectively to diverse tasks and datasets.


\noindent\textbf{Visual Comparison.}
Figure~\ref{fig:visual} presents qualitative comparisons across multiple prompts. With 8 NFE, TAFS-GRPO shows better semantic alignment than Flux.1-dev with 28 NFE, demonstrating effective few-step distillation. At the same 8 NFE, TAFS-GRPO also produces richer details and better aesthetics than Reward-Instruct.
The result of TAFS-GRPO reduces reward hacking compared to MixGRPO, which indicates the step-aware advantage integration precisely guides the early denoising steps, thereby mitigating the over-reliance on final-step rewards that drives the phenomenon.

\noindent\textbf{Flexibility of the Inference Step.}
As shown in Figure~\ref{fig:step_pickscore}, the TAFS-GRPO trained model achieves high-quality results with far fewer steps, significantly speeding up inference.
The model maintains high performance across a wide range of sampling steps (i.e., 4-8), providing flexibility for deployment under different latency constraints. Compared with other Flux-based distilled models (i.e., Hyper-Flux \cite{ren2024hyper}, Flux-Turbo-Alpha \cite{flux-turbo} and Flux.1-schnell \cite{blackforestlabs2024flux}), the TAFS-GRPO distilled model shows superior performance under different numbers of inference steps.
The consistent performance maintenance across varying step counts demonstrates the efficacy of our step-aware advantage integration strategy. Our method achieves optimal trade-offs between computational efficiency and generation quality, effectively addressing the core challenge in few-step flow-model sampling.
\subsection{Ablation Studies}

To validate the effectiveness of TAFS-GRPO, we perform ablation studies on both the temperature-annealed few-step sampling and the step-aware advantage integration mechanism, as shown in Table~\ref{tab:ablation}. 
For the ``w/o TAFS'' variant, we replace the temperature-annealed sampling in the TAFS-GRPO method with the same number of SDE sampling steps. Compared with TAFS-GRPO, the performance of this variant degrades on both in-domain and out-of-domain metrics, which verifies the usefulness of the temperature-annealed few-step sampling.
For the ``w/o Step Adv.'' variant, we replace the step-aware advantage integration with computing the group advantage of the final image rewards. This variant without the reward signal for each sampling step has inferior performance to the TAFS-GRPO method, which indicates that step-aware advantage integration can improve the performance of the TAFS-GRPO method.

\begin{table}[t]
  \centering
  \resizebox{\linewidth}{!}{
  \begin{tabular}{@{}lccc@{}}
    \toprule
    Method & Pick Score $\uparrow$ & CLIP Score $\uparrow$ & HPS-v2.1 $\uparrow$ \\
    \midrule
    TAFS-GRPO & 22.46 & 27.68 & 0.353 \\
    - w/o TAFS & 22.19 & 26.14 & 0.301 \\
    - w/o Step Adv. & 21.77 & 24.60 & 0.288 \\
    \bottomrule
  \end{tabular}
  }
  \caption{Ablation study on the components. TAFS means temperature-annealed few-step sampling method. Step Adv. means step-aware advantage integration mechanism.}
  \label{tab:ablation}
\end{table}

\textit{Due to the page limitation, additional ablation studies on the number of annealing transitions and the number of advantage-integration steps are provided in the Appendix.}

\section{Conclusion}
In this paper, we propose TAFS-GRPO, a method designed for efficient few-step training of flow-matching T2I models. We highlight the challenge of sampling efficiency and sparse reward when training flow matching models with RL. By integrating temperature-annealed few-step sampling into the GRPO framework, our approach improves sampling efficiency, which is a critical factor for online RL methods. To mitigate the sparse reward issue, we introduce a step-aware advantage integration mechanism, delivering more precise evaluation signals. As a result, TAFS-GRPO achieves accelerated few-step training on various flow matching models while enhancing alignment with human preferences on various prompt datasets and tasks. Notably, the few-step model trained by a single TAFS-GRPO process demonstrates robust performance across a wide range of sampling steps, offering flexibility for deployment under diverse latency constraints.

\bibliography{aaai2027}


\clearpage

\appendix

\section{Theoretical Analysis of TAFS}
\label{app:tafs-theory}

\subsection{A Stochasticity-Interpolated View of TAFS}

For theoretical analysis, we parameterize the rectified-flow path
by its normalized effective noise coordinate
$\sigma\in[0,1]$, where $\sigma=1$ corresponds to Gaussian noise
and $\sigma=0$ corresponds to clean data. Here, $\sigma$ denotes
the continuous noise coordinate associated with the scheduler,
rather than its discrete timestep index. We adopt the linear rectified-flow interpolation
\begin{equation}
z_{\sigma}
=
(1-\sigma)x_0+\sigma\epsilon,
\qquad
\epsilon\sim\mathcal{N}(0,I),
\label{eq:app-rf-bridge}
\end{equation}
and use the reverse velocity convention
$v=x_0-\epsilon$. Accordingly, the clean and noise endpoints
implied by a velocity prediction are
\begin{equation}
\widehat{x}_{0,\theta}(z_\sigma,\sigma,\mathbf c)
=
z_\sigma+\sigma
v_\theta(z_\sigma,\sigma,\mathbf c),
\label{eq:app-clean-endpoint}
\end{equation}
and
\begin{equation}
\widehat{\epsilon}_{\theta}(z_\sigma,\sigma,\mathbf c)
=
z_\sigma-(1-\sigma)
v_\theta(z_\sigma,\sigma,\mathbf c).
\label{eq:app-noise-endpoint}
\end{equation}

We consider an extended analytical schedule
$1=\sigma_0>\sigma_1>\cdots>\sigma_K=0.$
The finite training schedule corresponds to the positive noise
levels in this sequence. We append the clean endpoint
$\sigma_K=0$ solely to characterize the deterministic Euler limit;
it is not used as a stochastic re-noising transition during TAFS
training.

We introduce $\lambda$ solely as an analytical interpolation
parameter. For the positive noise levels used during training, the
TAFS sampler corresponds to the full-refresh case $\lambda=1$
under the linear rectified-flow parameterization
$\alpha_\sigma=1-\sigma$. In contrast, the annealing in TAFS refers
to the progressively decreasing noise levels $\{\sigma_k\}$, rather
than to a schedule over $\lambda$. For $\lambda\in[0,1]$, we define the following
stochasticity-interpolated transition:
\begin{equation}
\begin{split}
z_{\lambda}^{k+1}
={}&
(1-\sigma_{k+1})
\widehat{x}_{0,\theta}^{k}\\
&+
\sigma_{k+1}
\left(
\sqrt{1-\lambda^2}\,
\widehat{\epsilon}_{\theta}^{k}
+
\lambda\xi^k
\right),
\qquad
\xi^k\sim\mathcal{N}(0,I).
\end{split}
\label{eq:app-lambda-transition}
\end{equation}
When $\lambda=1$, Eq.~\eqref{eq:app-lambda-transition} reduces to
the full-refresh re-noising transition used by TAFS:
\begin{equation}
z_1^{k+1}
=
(1-\sigma_{k+1})\widehat{x}_{0,\theta}^{k}
+
\sigma_{k+1}\xi^k.
\label{eq:app-full-refresh}
\end{equation}

\paragraph{Proposition 1 (Oracle bridge-marginal preservation).}
Fix a prompt $\mathbf c$ and suppose that
\begin{equation}
z^k
=
(1-\sigma_k)x_0+\sigma_k\epsilon^k,
\qquad
\epsilon^k\mid x_0,\mathbf c
\sim\mathcal{N}(0,I).
\end{equation}
Assume that the endpoint predictions in
Eqs.~\eqref{eq:app-clean-endpoint} and
\eqref{eq:app-noise-endpoint} are replaced by the paired oracle
endpoints $x_0$ and $\epsilon^k$, respectively, and that
$\xi^k\sim\mathcal{N}(0,I)$ is conditionally independent of
$\epsilon^k$ given $(x_0,\mathbf c)$. Then, for every
$\lambda\in[0,1]$,
\begin{equation}
z_\lambda^{k+1}\mid x_0,\mathbf c
\sim
\mathcal{N}
\left(
(1-\sigma_{k+1})x_0,
\sigma_{k+1}^2I
\right).
\label{eq:app-marginal-preservation}
\end{equation}

\noindent\emph{Proof.}
Under the oracle endpoints,
Eq.~\eqref{eq:app-lambda-transition} becomes
\begin{equation}
z_\lambda^{k+1}
=
(1-\sigma_{k+1})x_0
+
\sigma_{k+1}\epsilon^{k+1},
\end{equation}
where
\begin{equation}
\epsilon^{k+1}
=
\sqrt{1-\lambda^2}\,\epsilon^k
+
\lambda\xi^k.
\end{equation}
Conditional independence gives
\begin{equation}
\mathbb{E}
[\epsilon^{k+1}\mid x_0,\mathbf c]
=0
\end{equation}
and
\begin{equation}
\operatorname{Cov}
(\epsilon^{k+1}\mid x_0,\mathbf c)
=
(1-\lambda^2)I+\lambda^2I
=
I.
\end{equation}
Since $\epsilon^k$ and $\xi^k$ are conditionally independent
Gaussian variables, their linear combination
$\epsilon^{k+1}$ is also conditionally Gaussian. Its conditional
mean is zero and its conditional covariance is $I$; hence
$\epsilon^{k+1}\mid x_0,\mathbf c\sim\mathcal N(0,I)$, 
which proves Eq.~\eqref{eq:app-marginal-preservation}.
\hfill$\square$

\paragraph{Corollary 1 (Zero-stochasticity Euler limit).}
At $\lambda=0$, substituting
Eqs.~\eqref{eq:app-clean-endpoint} and
\eqref{eq:app-noise-endpoint} into
Eq.~\eqref{eq:app-lambda-transition} yields
\begin{equation}
\begin{split}
z_0^{k+1}
&=
(1-\sigma_{k+1})
\left(z^k+\sigma_kv_\theta^k\right)\\
&\quad+
\sigma_{k+1}
\left(z^k-(1-\sigma_k)v_\theta^k\right)\\
&=
z^k+
(\sigma_k-\sigma_{k+1})v_\theta^k,
\end{split}
\label{eq:app-euler-limit}
\end{equation}
where
$v_\theta^k=v_\theta(z^k,\sigma_k,\mathbf c)$.
Equation~\eqref{eq:app-euler-limit} is the explicit-Euler update
of the reverse rectified-flow ODE
\begin{equation}
\frac{dz_\sigma}{d\sigma}
=
-v_\theta(z_\sigma,\sigma,\mathbf c).
\end{equation}

The interpolation parameter also has a direct correlation
interpretation. Under the assumptions of Proposition~1,
\[
\operatorname{Cov}
\left(
\epsilon^{k+1},\epsilon^k \mid x_0,\mathbf c
\right)
=
\sqrt{1-\lambda^2}I.
\]
Thus, $\lambda=1$ gives the independent full-refresh transition
used by TAFS, whereas $\lambda=0$ preserves the previous noise
and recovers the deterministic Euler update.


\paragraph{Remark.}
Proposition~1 is a per-time marginal statement under a paired
oracle-endpoint assumption. It does not imply that stochastic
TAFS and deterministic ODE sampling share the same conditional
transition kernels or the same joint trajectory distribution.
For a learned finite-step model, the marginal consistency is
approximate and depends on the endpoint-prediction and
one-step-transport errors.

\section{Experiment Setup Details}

\subsection{Training Settings}
For the composition image generation task, we jointly use GenEval, PickScore, CLIP score, and HPS-v2.1 as training rewards. For the human preference alignment task, we jointly use PickScore and CLIP score. We aggregate the reward scores using an equally weighted sum and then compute the group-normalized advantage from the aggregated reward. HPS-v2.1, ImageReward, and Unified Reward are used as out-of-domain evaluation metrics for the human preference alignment task.

We use the same optimization settings for all three base models, SD3.5-M, Flux.1-dev, and Flux.2-klein-4B. All models are trained at 512-pixel resolution with four one-step clean predictions and three reward-bearing annealing transitions. We optimize LoRA parameters with AdamW using a learning rate of \(3 \times 10^{-4}\), a weight decay of \(1 \times 10^{-4}\), and a group size of 48. For every model, LoRA uses rank $r=32$ and scaling parameter $\alpha=64$. The effective batch size per optimizer update is 48 trajectories, corresponding to 144 transition-level loss terms because each trajectory contributes three reward-bearing transitions. The distributed trajectory micro-batch size is 24 (3 per GPU across 8 GPUs).
We train SD3.5-M and Flux.1-dev using 8 NVIDIA H100 GPUs for 1500 iterations. The Flux.2-klein-4B run also uses 8 NVIDIA H100 GPUs, and Table~\ref{tab:distill} reports its checkpoint after 200 optimizer updates. All models are evaluated at 1024-pixel resolution using the metrics described below.

\subsection{Details on Evaluation Metrics}
For the composition image generation task, we use the GenEval score, pick score, CLIP score and human preference score as the reward function in training. We use the image reward and unified reward as the out-of-domain evaluation metrics.
For the human preference alignment task, PickScore and CLIP Score are used as both training rewards and in-domain evaluation metrics, while HPS-v2.1, ImageReward, and Unified Reward are used as out-of-domain evaluation metrics.

We introduce further details of the metrics used in the composition image generation task and human preference alignment task.

\noindent\textbf{GenEval Score.} 
GenEval \cite{ghosh2023geneval} score is an automated evaluation metric designed to assess text-to-image models on fine-grained, object-focused tasks such as object presence, counting, spatial relationships, color accuracy, and attribute binding. 

\noindent\textbf{Pick Score.} 
Pick score \cite{kirstain2023pick} is a CLIP-based scoring function trained on the Pick-a-Pic dataset, a large, open collection of real user preferences for text-to-image generation. Its primary purpose is to predict human preferences by evaluating how well a generated image aligns with a given text prompt, which maximizes the probability that a preferred image is ranked higher than a non-preferred one. 

\noindent\textbf{CLIP Score.}
The CLIP Score \cite{hessel2021clipscore} is a reference-free evaluation metric that quantifies the semantic alignment between a generated image and its corresponding text prompt. It leverages the pre-trained CLIP model, which projects both images and text into a shared embedding space. The score is computed as the cosine similarity between the image embedding and the text embedding, measuring how closely the visual content matches the textual description in a high-level semantic sense.

\noindent\textbf{HPS-v2.1 Score.}
Human Preference Score v2.1 \cite{wu2023human} is a metric designed to evaluate the alignment of text-to-image generative models with human aesthetic preferences. 

\noindent\textbf{ImageReward}
ImageReward \cite{xu2023imagereward} is a general-purpose text-to-image human preference reward model designed to effectively encode human preferences for images generated from text prompts.  It was trained on a large-scale dataset of 137,000 expert comparisons collected through a systematic annotation pipeline involving rating and ranking. 

\noindent\textbf{Unified Reward.}
The Unified Reward Model \cite{wang2025unified} is designed to assess both multimodal understanding and generation tasks. It overcomes the limitations of traditional, task-specific reward models by leveraging joint learning across diverse visual tasks, which creates a synergistic effect where improvements in one domain (e.g., image understanding) enhance performance in another (e.g., image assessment).

\begin{figure*}[!t]
    \centering
    \includegraphics[width=0.90\linewidth]{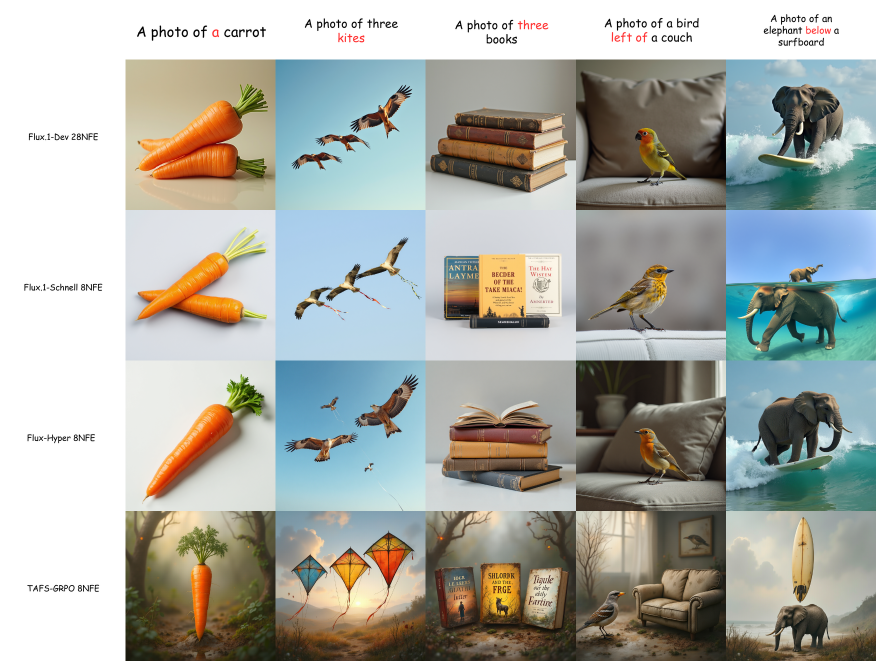}
    \caption{Qualitative comparison of generations on the composition image generation task from 28 sampling steps of Flux.1-dev, and 8 sampling steps of Flux.1-Schnell, Hyper-Flux, and the proposed TAFS-GRPO.}
    \label{fig:visual_genval}
\end{figure*}

\section{Extended Experimental Results}

\subsection{Additional Qualitative Results}
Figure \ref{fig:visual_genval} shows the qualitatively compare TAFS-GRPO with Flux.1-dev, Flux.1-Schnell and Hyper-Flux on the composition image generation task. The results of TAFS-GRPO show significant improvements in object-focused tasks such as object presence, counting, and spatial relationships.

\subsection{Ablation Study on the Number of Annealing Transitions}
As shown in Table~\ref{tab:num_step}, the performance of TAFS-GRPO generally improves as the number of stochastic annealing transitions increases, which validates the efficacy of the proposed framework. To balance training speed with performance, we select 3 annealing transitions, corresponding to 4 one-step clean predictions, for the experiments.

\begin{table}[h]
  \centering
  \resizebox{\linewidth}{!}{
  \begin{tabular}{@{}ccccc@{}}
    \toprule
    Anneal. Trans. & Time $\downarrow$ & Pick Score $\uparrow$ & CLIP Score $\uparrow$ & HPS-v2.1 $\uparrow$ \\
    \midrule
    1 & 47  & 21.76 & 27.14 & 0.296	\\
    3 & 116 & 22.46 & 27.68 & 0.353 \\
    5 & 185 & 22.53 & 27.31 & 0.363 \\
    \bottomrule
  \end{tabular}
  }
  \caption{Ablation study on the number of stochastic annealing transitions in TAFS-GRPO training. A setting of $K$ transitions corresponds to $K+1$ one-step clean predictions, $K+1$ old-policy evaluations, and $K$ policy evaluations for likelihood-ratio recomputation. The Time column reports the wall-clock time in seconds per training iteration.}
  \label{tab:num_step}
\end{table}

\subsection{Ablation Study on the Number of Advantage Integration Steps}
To validate the effectiveness of step-aware advantage integration, we perform ablation studies on various numbers of advantage integration steps shown in Table~\ref{tab:num_adv}. The performance of TAFS-GRPO shows consistent improvement with an increasing number of advantage integration steps, providing compelling evidence that the step-aware advantage integration successfully directs reward signals to precisely optimize the model. As indicated in the Time column, the step-aware advantage integration introduces minimal computational overhead and does not significantly increase training time.

\begin{table}[h]
  \centering
  \resizebox{\linewidth}{!}{
  \begin{tabular}{@{}ccccc@{}}
    \toprule
    N$_{Adv.}$ & Time $\downarrow$ & Pick Score $\uparrow$ & CLIP Score $\uparrow$ & HPS-v2.1 $\uparrow$ \\
    \midrule
    1 & 107 & 21.77 & 24.60 & 0.288 \\
    2 & 112 & 22.02 & 26.17 & 0.301 \\
    3 & 116 & 22.46 & 27.68 & 0.353 \\
    \bottomrule
  \end{tabular}
  }
  \caption{Ablation study on the number of advantage integration steps in TAFS-GRPO training. N$_{\mathrm{Adv.}}$ denotes the number of sampling steps used for advantage integration. $N_{\mathrm{Adv.}}=1$ uses only the final image reward and corresponds to the ``w/o Step Adv.'' baseline in Table~\ref{tab:ablation}. The number of annealing transitions is set to 3. The Time column denotes the time in seconds per training iteration.}
  \label{tab:num_adv}
\end{table}

\subsection{Reward Model Sensitivity Analysis}

\begin{table}[h]
\centering
\resizebox{\linewidth}{!}{
\begin{tabular}{lcccc}
\toprule
Reward Model & Pick Score $\uparrow$ & CLIP $\uparrow$ & HPS-v2.1 $\uparrow$ & Unified R $\uparrow$ \\
\midrule
CLIP-only     & 22.01 & 27.92 & 0.312 & 3.421 \\
PickScore-only     & 22.58 & 26.74 & 0.298 & 3.398 \\
Combined      & 22.46 & 27.68 & 0.353 & 3.511 \\
\bottomrule
\end{tabular}
}
\caption{Experiment on reward sensitivity with Flux.1-dev base model and tested under 8 inference steps.}
\label{tab:reward_sensitivity}
\end{table}

We conducted a sensitivity analysis to evaluate the impact of different reward model configurations during training. The results in Table~\ref{tab:reward_sensitivity} indicate that using a single reward model leads to noticeable performance degradation in certain out-of-domain metrics. In contrast, the combined reward model achieved the best overall balance across both in-domain and out-of-domain evaluation metrics. These findings suggest that multi-reward integration enhances model robustness and effectively mitigates overfitting to any single metric.

\subsection{Analysis of Intermediate Reward Validity}
\label{sec:intermediate_validity}

\begin{table}[!b]
    \centering
    \small
    \begin{tabular}{lcc}
    \toprule
    {Step Reward} & {Pick Score ($r$)} & {CLIP Score ($r$)} \\
    \midrule
    $R^2$ with $R^4$ & 0.68 & 0.72 \\
    $R^3$ with $R^4$ & 0.84 & 0.88 \\
    \bottomrule
    \end{tabular}
    \caption{Correlation between intermediate step rewards and final step rewards.}
    \label{tab:correlation}
\end{table}

A key premise of TAFS-GRPO is that the one-step clean estimates $x_0^n$ derived from noisy states $x_{t_n}^{n-1}$ contain sufficient semantic content to provide valid reward signals. To validate this, we computed the Pearson correlation coefficient between the reward scores of the intermediate estimates and the final generated image across 1,000 training prompts. As shown in Table \ref{tab:correlation}, even the early step reward $R^2$ estimates show a strong positive correlation ($r > 0.65$) with the final quality. This indicates that although early estimates may lack high-frequency details, they successfully capture the global semantic structure (e.g., object presence, composition) required by reward models like CLIP and PickScore. This validates the use of dense supervision, as the signal provided by early steps guide the model in a direction consistent with the final objective.

\end{document}